\documentclass{article}
\usepackage[dblblindworkshop, final]{neurips_2025}

\usepackage[utf8]{inputenc}
\usepackage[T1]{fontenc}
\usepackage{hyperref}
\usepackage{url}
\usepackage{booktabs}
\usepackage{amsfonts}
\usepackage{nicefrac}
\usepackage{microtype}
\usepackage{xcolor}
\usepackage{array}
\usepackage{graphicx}
\usepackage{amsmath}
\usepackage{amssymb}
\usepackage{float}
\usepackage{wrapfig}
\usepackage{caption}
\usepackage{subcaption}
\usepackage{lipsum}
\usepackage{tikz}
\usepackage{graphicx} 
\usetikzlibrary{shadows,arrows,positioning,calc}
\pgfdeclarelayer{background}
\pgfdeclarelayer{foreground}
\pgfsetlayers{background,main,foreground}

\title{Machine Learning and Multi-source Remote Sensing
in Forest Aboveground Biomass Estimation: A Review}
\workshoptitle{Tackling Climate Change with Machine Learning}

\author{%
Autumn Nguyen \\
  Mount Holyoke College \\
  Computer Science \\
  South Hadley, MA, United States \\
\texttt{autumn.yngoc@gmail.com} 
  \And
  Sulagna Saha \\
  Mount Holyoke College \\
  Computer Science \\
  South Hadley, MA, United States \\
\texttt{saha23s@mtholyoke.edu} 
}

\begin{document}
\maketitle

\begin{abstract}
Quantifying forest aboveground biomass (AGB) is crucial for informing decisions and policies that will protect the planet. Machine learning (ML) and remote sensing (RS) techniques have been used to do this task more effectively, yet there lacks a systematic review on the most recent working combinations of ML methods and multiple RS sources, especially with the consideration of the forests' ecological characteristics. This study systematically analyzed 25 papers that met strict inclusion criteria from over 80 related studies, identifying all ML methods and combinations of RS data used. Random Forest had the most frequent appearance (88\% of studies), while Extreme Gradient Boosting showed superior performance in 75\% of the studies in which it was compared with other methods. Sentinel-1 emerged as the most utilized remote sensing source, with multi-sensor approaches (e.g., Sentinel-1, Sentinel-2, and LiDAR) proving especially effective. Our findings provide grounds for recommending which sensing sources, variables, and methods to consider using when integrating ML and RS for forest AGB estimation.
\end{abstract}

\section{Introduction}
The main driver of the increasing global deforestation is because forests
are mainly valued in terms of their economic value, such as how much timber or area of land they can provide, rather than on how
much they help regulate climate \cite{seymour2017forests}. To quantify how keeping certain areas of forests can help the climate, we need to quantify their carbon stock. The best way to measure the amount of carbon sequestered in a forest is to do so with direct field measurements \cite{Cunningham2011}, but since manually collecting field measurements at large scale is too costly, RS has been utilized. RS can fly over large areas of forests to capture information, such as tree density, vegetation cover, and 3D structures, with minimal disturbance, and these data can be put into forestry allometric equations to calculate AGB. Combining multiple sources remote sensing is promising because it allows the capabilities of one source to compensate for the limitations of the other sources, as we summarized in Table~\ref{table1}. There have been studies aiming to summarize existing work in this area. Ouaknine et al. \cite{ouaknine2024openforest} provided the comprehensive list of open forest monitoring datasets. Sun and Liu \cite{sun2020review} reviewed fundamental estimation methods, but only for studies in China, and they did not review any ML methods. Rolnick et al. \cite{tacklingCCwML} provided a broad overview of ML in climate change, but with limited focus on forest carbon. Hamedianfar et al. \cite{deeplearningreview} detailed deep learning methods for forest inventory, but no common non-DL methods like RF; plus this highly technical approach may be inaccessible to a broad audience. Matiza et al. \cite{matizareview} reviewed ML and RS approaches for carbon storage, but they did not analyze the specific combinations of data sources or forest characteristics. Our study addresses those gaps by reviewing studies done around the world, focusing on forest AGB estimation with ML and combinations of RS sources (i.e. multi-source RS), and communicating the results in an accessible way to people who may not have deep expertise in those areas.

\section{Methods}

The papers in our review were drawn from these \textbf{search terms}: “(estimation OR estimating OR "machine learning") (multisource OR multi-source OR multisensor OR multi-sensor) forest carbon biomass map”. The aim was to find papers that surely had a Machine Learning component, used a combination of different sources of remote sensing data, and for the purpose of mapping carbon stick or biomass of forests. We retrieved the papers from this search into a database using the public API from \cite{ankit}, specifically the google\_scholar\_internal function. The 25 papers that we drew quantitative results from were the first 25 that satisfied five of our \textbf{inclusion criteria}: (1) full paper accessible to us (so most are open access articles, since we are college students with almost no subscriptions to any journals), (2) used ML in the study, (3) used multiple sources of remote sensing data, (4) had the end goal being estimation forest carbon, whether it was AGB or BGB or soil carbon, and (5) was written in the recent 10 years (2014-2024).

The following \textbf{data} was collected for the \textbf{quantitative database}: 1) All the remote sensing sources that the study took data from. 2) All the ML methods used in the study. 3) If the study used multiple ML methods for comparison of performance on the same task, or for each method to be used on a different task. 4) If the study used multiple ML methods for comparison, which method(s) were found to have the best performance? Since there are many different ways to define "best", we just included the methods that were explicitly mentioned in the abstract or conclusion with a keyword "best", or "highest" for metrics like $R^2$ for accuracy, or "lowest" for metrics like RMSE or uncertainty. 5) Any limitations or future steps thoroughly explained. 6) The ultimate task, such as AGB map, BGB map, general biomass map, multi-scale biomass maps, uncertainty estimation, etc. 7) The location(s) of the studied forests. 8) The types or characteristics or dominant species of the forests. The forest type categories we used in our review are not at all mutually exclusive or deterministic — they are meant to facilitate readers in identifying the papers that work on forests of similar types to their interest. Since the words people used to describe their forests varied widely between papers, we did our best to identify the common terms used across papers, and refer to a few sources (\cite{xu2022global, arcgis}) to determine the forest types of the papers that did not use exactly those common terms (so we related their terms to the common terms), and of the papers that did not have any terms about the type of their forest (for those, we used the geographical latitude and longitude of the area to determine the type based on the external sources cited above). 9) The scale of the study: region, country, or global. Python libraries, namely Pandas, Matplotlib, Seaborn, and NumPy, were used to manipulate and visualize data.

\section{Results \& Discussions}
We created a \href{https://itsautumn.notion.site/10a0d405e6518047b073ddd00c71dc65?v=12b0d405e6518078be9b000cbf6ccde5&pvs=4}{interactive database}\footnote{\url{https://itsautumn.notion.site/10a0d405e6518047b073ddd00c71dc65?v=12b0d405e6518078be9b000cbf6ccde5&pvs=4}} which everyone can filter by forest types, data sources, ML methods, or any keywords they want to find relevant papers that we reviewed. A summary table and the abbreviations of ML methods and data sources can be found in the Appendix.

\begin{figure}[H]
    \centering
    \includegraphics[width=0.70\linewidth]{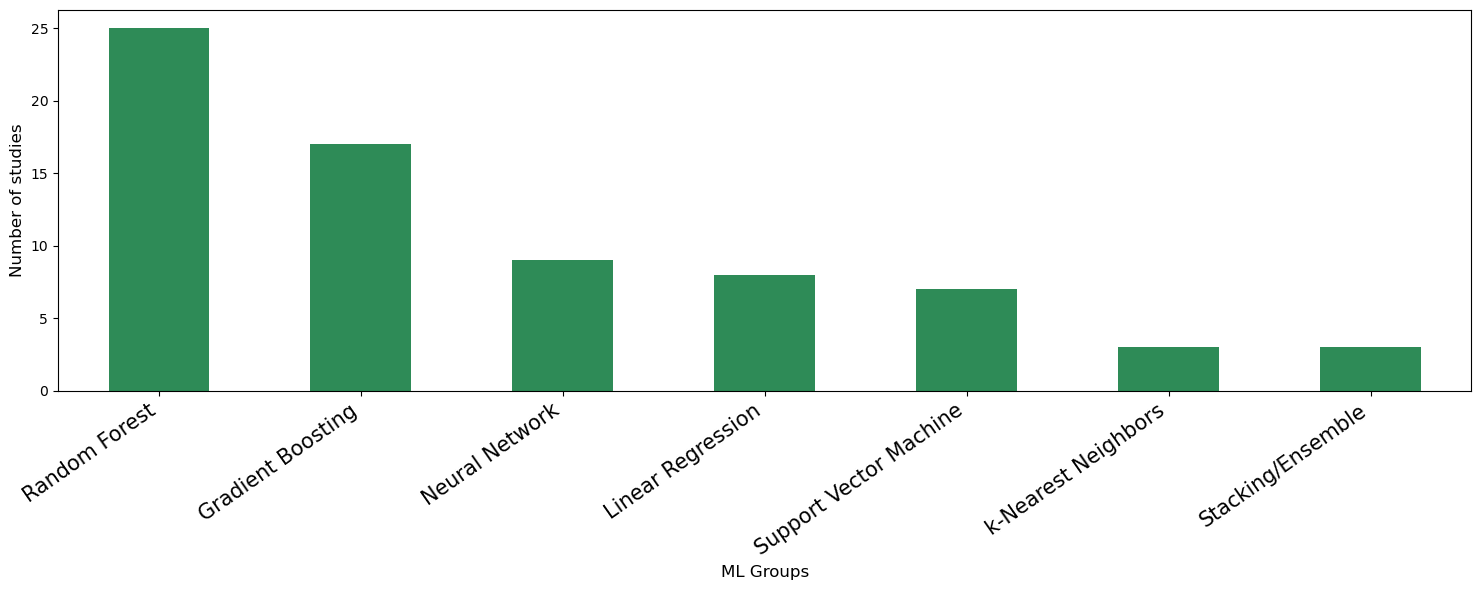}
    \caption{Frequency of ML methods by Groups}
    \label{ml_groups}
\end{figure}

Most of the ML methods had roughly similar appearance frequency (see Appendix), except for Random Forest, which was used in around 88\% of the studies -- as the model for the end task—AGB estimation and sometimes as the model for other intermediate tasks in the data processing pipeline. We put the methods into groups in Figure \ref{ml_groups} to see the trend in a bigger picture. Random Forests are still the most commonly used methods, and most of the methods found to have the best performance fall into the three most frequently used groups: Random Forest (RF, QRF, RRF), Gradient Boosting (XGB, LGBM, CatBoost), and Neural Network (CNN, BayesResNN). 11 out of 25 studies compared multiple ML methods for the AGB estimation task, and found the model(s) that performed best. RF was part of all the studies that compared multiple MLs, but was only found to be best in 4, whereas XGB was only used in 4 studies, but was found to perform the best in 3. For instance, \cite{Li2020} found that XGB had best estimations of AGB in high and low range values, while XGB, RF, LR performed similarly in medium range, so XGB also improved the overestimation-underestimation issue. 

The most frequently used data source was Sentinel-1, followed by Sentinel-2, ALOS-PALSAR, Landsat, MODIS, GLAS/ICESat LiDAR, and GEDI LiDAR. However, since GLAS/ICESat and GEDI are both spaceborne LiDAR, we can also say that spaceborne LiDARs were the most frequently used source. In 2022, Sentinel-1 only came at the 7th, ALOS-PALSAR the 13th, while Landsat sat at the top of the frequency rank of sensors in \cite{matizareview}.

\begin{figure}[H]
    \centering
    \includegraphics[width=0.7\linewidth]{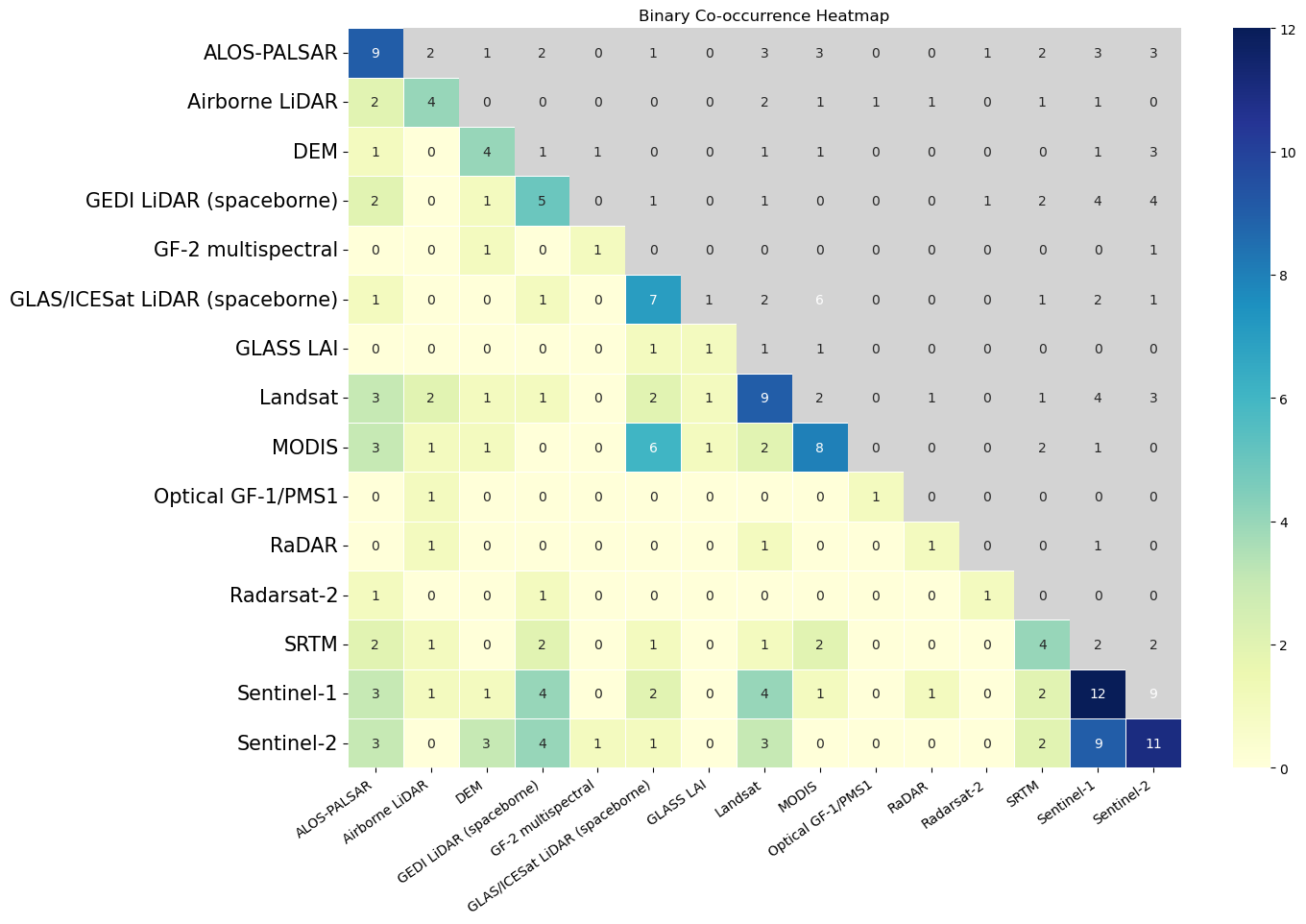}
    \caption{Heatmap: binary co-occurrence of two data sources}
    \label{sources_heatmap}
\end{figure}

The heatmap \ref{sources_heatmap} shows that Sentinel-1, Sentinel-2, and spaceborne LiDAR (GEDI) were most often used together. They made a combination of passive optical, active optical, and radar, complementing each other’s strengths and limitations. LiDAR had very limited availability and high costs, so when it was not available, combinations of passive optical and radar also used well together quite often. For example, Landsat and Sentinel-1, or MODIS and Sentinel-2, or MODIS and PALSAR. Nonetheless, some of those studies that used only passive optical and radar sensors faced a common issue of saturation, and LiDAR was usually the recommended solution to that issue. Another observation was that when a study used Sentinel-2, they’d likely also include LiDAR or DEM. This is likely because Sentinel-2 is a passive optical sensor with no ability to infer canopy height, an important variable in estimating AGB, and LiDAR or DEM can provide that information. 

\textbf{Feature selection} was found to be a critical factor influencing the models’ performances in many studies. A key finding from the literature is that a well-executed variable selection process can significantly enhance model accuracy. For instance, Li et al. \cite{Li2019} demonstrated that feature selection substantially improved the performance of all their models, with XGBoost showing the most significant gains. Similarly, Huang et al. \cite{Huang2022} used the Least Absolute Shrinkage and Selection Operator (Lasso) to reduce over 30 initial numerical parameters to just seven.
However, there is a contrasting perspective in the literature. Some studies, such as that by Li et al. \cite{Li2020}, opted against feature selection. Their decision was based on two main arguments. First, their datasets had a limited number of variables, and they were concerned that multiple rounds of selection would eliminate crucial data, particularly from Sentinel sensor variables. Second, they argued that their chosen models, specifically Random Forest (RF) and XGBoost, were inherently robust against noisy variables. They noted that RF is generally unaffected by noisy predictors, while XGBoost's regularization objective helps to restrain their influence. Since this was somewhat contrasting with how many studies using RF and XGB did perform feature selection and noted the improved performance, it would be an interesting topic to look into the relationship between feature selection and choice of ML models. 

\textbf{3D structural data} from Shuttle Radar Topography Mission or other digital elevation models seems to be frequently used in the studies that didn’t have LiDAR data, when it was used, it was usually one of the the most important predictor variables.  \cite{huang2023estimating} had Landsat OLI and Sentinel-2 as the two main remote sensors, with the addition of DEM data, and found that DEM was the most important variable. \cite{Huang2022} also had optical sensors and radar sensor as Sentinel-2B, Sentinel-2A, Sentinel 1A, with the inclusion of DEM data.

There are also \textbf{variables not from remote sensors} that were found to be critical as well. \textbf{Phenological characteristics}, the seasonal patterns and timing of biological events of different forest types and different tree species, was found to be a valuable one. When \cite{Zhang2023} inputed phenological variables into their model, they achieved a higher R-squared result. Although their study area had a specific dominant forest type and dominant species, an implication of their work was that incorporating data about phenological characteristics and dominant species significantly improved the accuracy of AGB estimation. Phenological characteristics were also used to help extracting the distribution information of their study subject (larch trees) in \cite{Hong2023}. In addition, the close relationship between phenological data and \textbf{time} were a major advantage for AGB estimations. Forests’ carbon flux varies over seasons, but the commonly used spectral variables from optical sensors only reflected the state of the forests at one point in time. Therefore, the AGB estimation based solely on those variables couldn’t be scaled temporally \cite{Zhang2023}. With phenological variables in play, we can create what \cite{Zhang2023} called time-consistent AGB models. In \cite{Hong2023}’s study, their LSTM model did well in the last stage of their study pipeline, which was extrapolating biomass components at the regional scale. It is reasonable because LSTM is a type of recurrent neural networks that is specialized in working with time-series and sequential data, and time is an important indicator in the phenological data they used. When \cite{Hong2023} compared LSTM with RF for this task, they found that the LSTM model was less prone to underestimation of biomass, and this characteristic became more obvious when the sample unit biomass was increased.

Another insight was that \textbf{different ML algorithms may be suitable for different stages of a forest carbon mapping pipeline}
In mapping AGB in alpine regions of Yunnan, \cite{Zhang2023} used three different ML methods through their pipeline: logistic regression to extract phenological parameters from Landsat and work with MCCDC; SVM to take in in phenological parameters and classify forest dominant tree groups; and RF to take in forest dominant tree groups mapping and create AGB map for the region. \cite{Hong2023} compared RF and MLR for creating Plot-Scale Biomass Component Estimation Model; used SVM for the extraction of Larch Distribution Information on the Basis of Vegetation Phenology Characteristics; and compared RF and LSTM for the extrapolation of Biomass Components at the Regional Scale. \cite{xi2023carbon} used an optimized RF regressor to calculate early estimates of carbon storage at the canopy scale in the footprints of ICESat-2/ATLAS LiDAR data; and used a deep neural network to create regional-scale carbon storage maps from those early LiDAR estimates and Landsat data. 

\section{Conclusion \& Future Steps}

This review highlights the machine learning methods and the remote sensing sources and combinations with the highest usage frequency and performance. Our recommendation for future studies on estimating forest AGB is to, in terms of remote sensing data sources,  combine multiple sources of remote sensing data, at least passive optical and radar optical, such as Sentinel-1 with Sentinel-2 or MODIS with ALOS-PALSAR, to address coverage and saturation limitations. In addition, including data that can indicate the forest's 3D structure, like from DEM or active optical sensors, can enhance accuracy and mitigate the overestimation-underestimation problem. In terms of ML methods, Random Forest is a great baseline method due to its long history of reliability, but it may also be worth trying other methods that had proven success recently, such as Extreme Gradient Boosting or CNN. We also emphasize the importance of feature selection and ensuring the spatial heterogeneity of sample plots to improve model performance. Additionally, rather than just using one ML method, different ML methods can be leveraged at various stages of the data processing pipeline. Future work should also take into account the types, phenological characteristics, and dominant species of the forests in building estimation models.

\begin{ack}
We are grateful for Professor \textbf{Alyx Burns}, who has met with us every week throughout Fall 2024 to advise us on the whole process, from reading papers to compiling the database to visualizing results. We are also deeply thankful for Dr. \textbf{Sreedath Panat}, Dr. \textbf{David Dao}, and Dr. \textbf{Björn Lütjens} who have given us not only advice and insights, but also great encouragement. 

No funding was received for this work.

\end{ack}
\newpage
\bibliographystyle{plain}
\bibliography{main}

\newpage
\section{Appendix}
\begin{table}[H]

\centering
{

\begin{tabular}{| l  |>{\raggedright\arraybackslash}p{0.2\linewidth}|>{\raggedright\arraybackslash}p{0.2\linewidth}|>{\raggedright\arraybackslash}p{0.2\linewidth}|}
\hline
 & Passive optical & Active optical (LiDAR) & Radar (SAR) \\
\hline
Sensors & Sentinel-2

MODIS

Landsat & GLAS (ICESat) & Sentinel-1

ALOS-PALSAR \\
\hline
Capabilities & Widely and freely available

High resolution & Able to measure 3D structural data

Can work at nighttime

High resolution & Can penetrate dense canopies and clouds

Can work at nighttime \\
\hline
Limitations & Cannot work at nighttime

Cannot penetrate dense canopies or clouds & Expensive

Available in only certain small areas 

Cannot penetrate clouds & Lower resolution than optical

Saturation issues \\
\hline

\end{tabular}
}
\caption{Capabilities and limitations of most common categories of remote sensor modalities}
\label{table1} 
\end{table}

\subsection{Abbreviations of ML methods}

\begin{itemize}
    \item Random Forest (RF)
    \item Quantile Random Forest (QRF)
    \item Regularized Random Forest (RRF)
    \item Extremely Randomized Trees (ERT)
    \item Gradient Tree Boosting (GTB)
    \item Gradient-Boosted Regression Tree (GBRT)
    \item Boosted Regression Tree (BRT)
    \item Gradient Boosting Machine (GBM)
    \item Light Gradient Boosting Machine (LGBM)
    \item Stochastic Gradient Boosting (SGB)
    \item Extreme Gradient Boosting (XGB)
    \item Categorical Boosting (CatBoost)
    \item Linear Regression (LR)
    \item Multi-Linear Regression (MLR)
    \item Stepwise Linear Regression (StepwiseLR)
    \item Multivariate adaptive regression splines (MARS)

    \item Random Forest with Stacking Algorithm (RFStacking)
    \item Cubist Regression Tree Ensemble (CubistRTEns)
    \item Stacked Ensemble for RF and boosting algorithms 
    \item Bayesian Regularization Neural Network (BayesRegNN)

\end{itemize}

\subsection{Groupings of ML methods}
The ML methods were grouped as follows:

\begin{itemize}
    \item 'Random Forest': ['RF', 'RRF', 'QRF', 'ERT'],
    \item 'Gradient Boosting': ['GTB', 'GBM', 'GBRT', 'BRT', 'LGBM', 'SGB', 'XGB', 'CatBoost'],
    \item 'Linear Regression': ['MLR', 'LR', 'Stepwise LR', 'MARS'],
    \item 'Neural Networks': ['LSTM', 'QRNN', 'CNN', 'ANN', 'BayesRegNN', 'Keras'],
    \item 'Support Vector Machines': ['SVM', 'SVR'],
    \item 'Stacking/Ensembles': ['RFStacking', 'StackedEnsemble'],
    \item 'Cubist': ['CubistRTEns'],
    \item 'k-NN': ['kNN']
\end{itemize}

\subsection{Summary table}
\begin{table}[H]

\centering

\begin{tabular}{|>{\raggedright\arraybackslash}m{0.2\linewidth}|>{\raggedright\arraybackslash}m{0.5\linewidth}|>{\raggedright\arraybackslash}m{0.29\linewidth}|} \hline 
Study & Data sources & ML methods used \\ \hline 
\cite{fararoda2021improving} & ALOS-PALSAR, DEM, MODIS & RF, kNN \\ \hline 
\cite{zhu2020estimation} & ALOS-PALSAR, Landsat & MLR, RF \\ \hline 
\cite{Hong2023} & Airborne LiDAR, Optical GF-1/PMS1 & LSTM, MLR, RF, SVM \\ \hline 
\cite{wang2022integrating} & ALOS-PALSAR, GLAS/ICESat LiDAR (spaceborne), MODIS & LR, QRNN, RF, SVM, Stepwise LR \\ \hline 
\cite{zhang2019estimating} & GLAS/ICESat LiDAR (spaceborne), Landsat, MODIS & CubistRegrTree \\ \hline 
\cite{zhangcnn} & ALOS-PALSAR, Sentinel-1, Sentinel-2 & CNN, Keras, MLR, RF, SVM \\ \hline 
\cite{tang2022estimation} & GLAS/ICESat LiDAR (spaceborne), MODIS, SRTM & CatBoost, GBM, LGBM, RF, XGB \\ \hline 
\cite{ehlers2022mapping} & Airborne LiDAR, Landsat, RaDAR, Sentinel-1 & RF \\ \hline 
\cite{hu2020mapping} & GLAS/ICESat LiDAR (spaceborne), MODIS & RF \\ \hline 
\cite{huang2023estimating} & DEM, Landsat, Sentinel-2 & BayesRegNN, GBM, QRF, RF, RRF, kNN \\ \hline 
\cite{ometto2023biomass} & ALOS-PALSAR, Airborne LiDAR, MODIS, SRTM & RF \\ \hline 
\cite{ghosh2018aboveground} & Sentinel-1, Sentinel-2 & RF, SGB \\ \hline 
\cite{bispo2020woody} & ALOS-PALSAR, Airborne LiDAR, Landsat & RF \\ \hline 
\cite{potzschner2022ecoregion} & GLAS/ICESat LiDAR (spaceborne), MODIS, Sentinel-1 & GBM \\ \hline 
\cite{singh2024optimising} & Sentinel-1, Sentinel-2 & ANN, RF, SVM \\ \hline 
\cite{ronoud2021multi} & Landsat, Sentinel-1, Sentinel-2 & MLR, RF, SVR, kNN \\ \hline 
\cite{pakistan} & DEM, GEDI LiDAR (spaceborne), Sentinel-1, Sentinel-2 & CatBoost, GTB, LGBM, RF, XGB \\ \hline 
\cite{chen2022improved} & GEDI LiDAR (spaceborne), GLAS/ICESat LiDAR (spaceborne), Sentinel-1, Sentinel-2 & RF \\ \hline 
\cite{musthafa2022improving} & ALOS-PALSAR, GEDI LiDAR (spaceborne), Radarsat-2 & RF \\ \hline 
\cite{li2021novel} & DEM, GF-2 multispectral, Sentinel-2 & RFStacking \\ \hline 
\cite{indiaxgb} & GEDI LiDAR (spaceborne), SRTM, Sentinel-1, Sentinel-2 & BRT, RF, XGB \\ \hline 
\cite{zhang2020evaluation} & GLAS/ICESat LiDAR (spaceborne), GLASS LAI, Landsat, MODIS & ANN, CatBoost, ERT, GBRT, MARS, RF, SGB, SVR \\ \hline 
\cite{ghosh2022predicting} & ALOS-PALSAR, GEDI LiDAR (spaceborne), Landsat, SRTM, Sentinel-1, Sentinel-2 & RF \\ \hline 
\cite{Li2020} & Landsat, Sentinel-1 & LR, RF, XGB \\ \hline

\end{tabular}
\caption{Data sources and ML methods used in selected studies.}
\label{Table2}

\end{table}

\subsection{Visualizations of Quantitative Results}
\begin{figure}[H]
    \centering
    \includegraphics[width=0.99\linewidth]{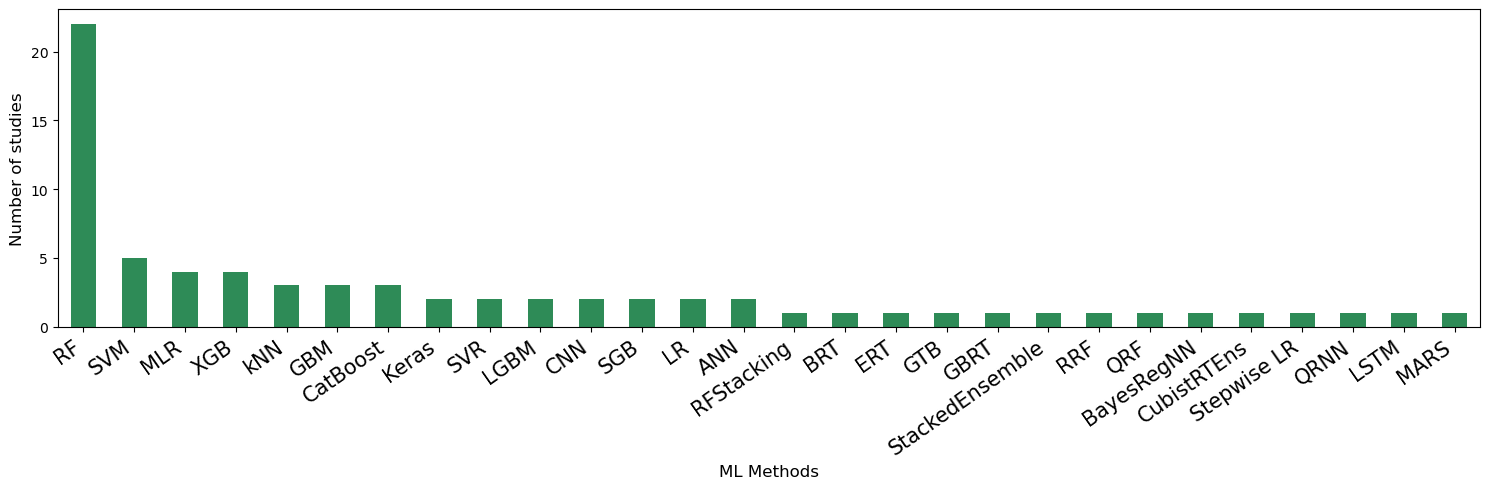}
    \caption{Frequency of ML methods used}
    \label{ml_all}
\end{figure}

\begin{figure}[H]
    \centering
    \includegraphics[width=0.85\linewidth]{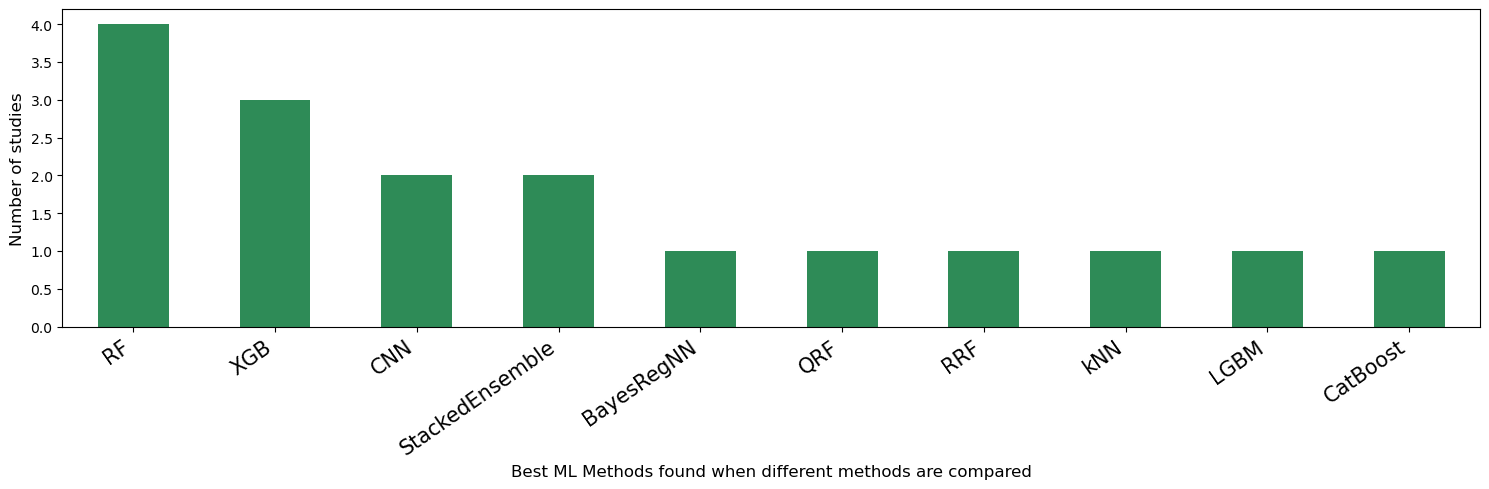}
    \caption{Best ML methods found in studies that compared multiple methods}
    \label{ml_best}
\end{figure}

\begin{figure}[H]
    \centering
    \includegraphics[width=0.49\linewidth]{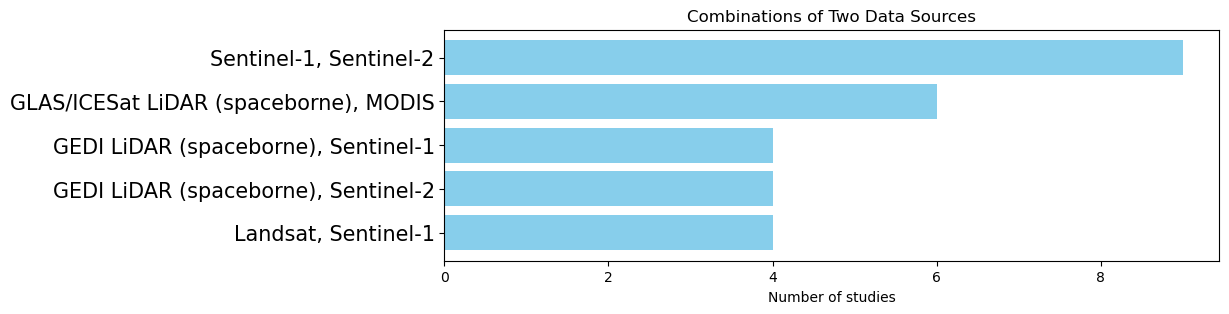}
    \includegraphics[width=0.49\linewidth]{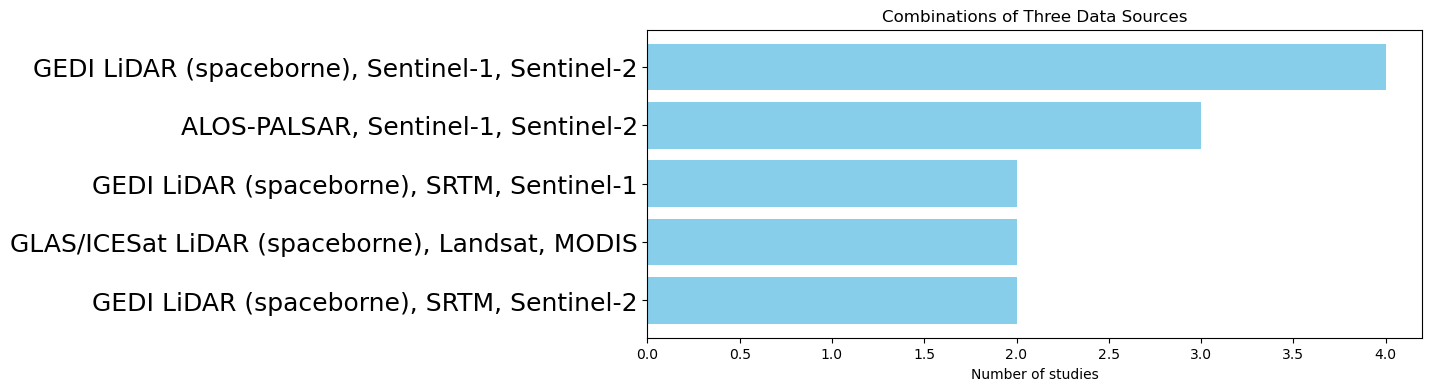}
    \caption{Most common combinations of data sources}
    \label{sources_duos_trios}
\end{figure}

\begin{figure}[H]
    \centering
    \includegraphics[width=0.8\linewidth]{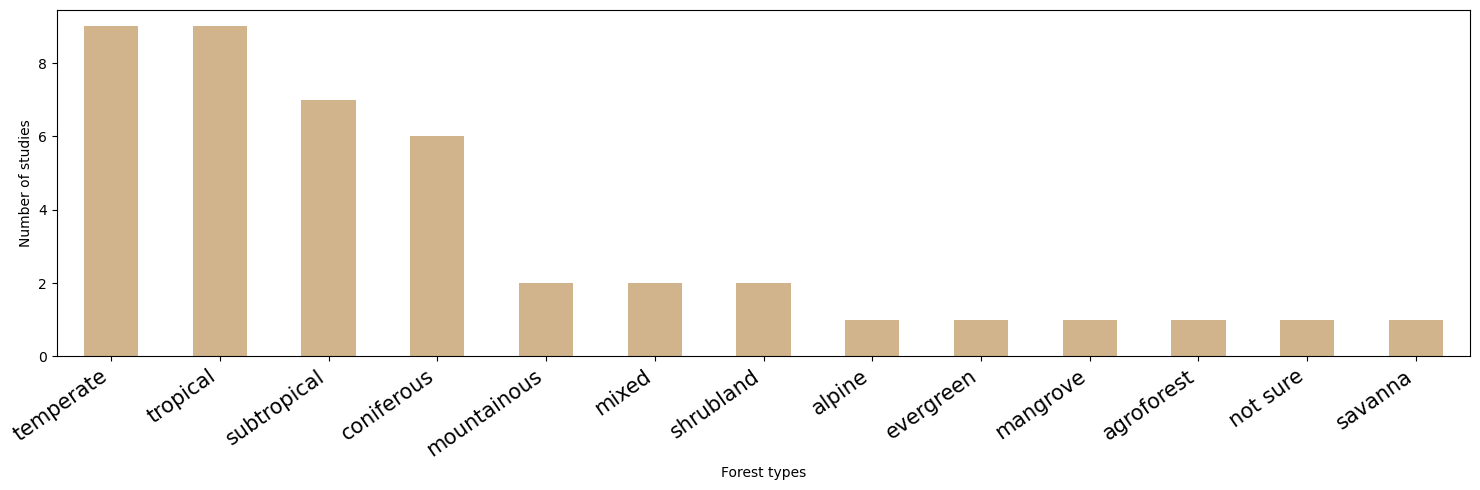}
    \caption{Frequency of forest types studied}
    \label{forests_type}
\end{figure}

\begin{figure}[H]
    \centering
    \includegraphics[width=0.8\linewidth]{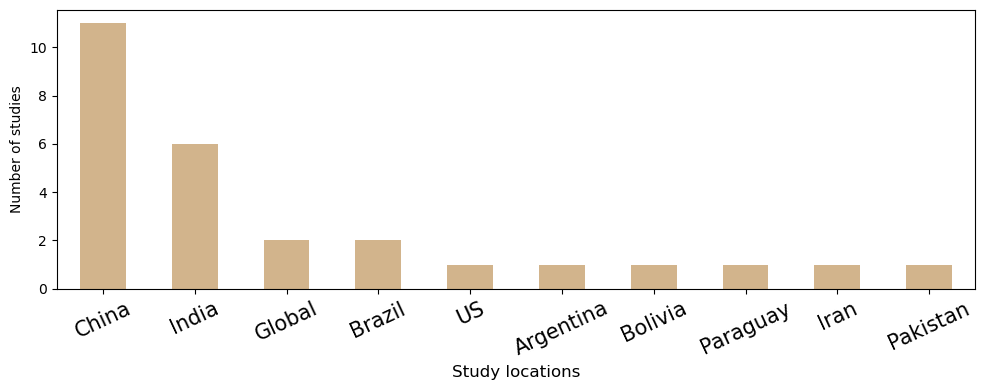}
    \caption{Frequency of geographical locations of the studied forests}
    \label{fig:locations}
\end{figure}
China made up around half of the locations of the studies, making it the most frequently appeared country in Figure \ref{fig:locations}. Since none of the terms were China-related, and all papers were selected from the first-appeared results on Google Scholar rather than from related papers, this may point to some interesting geographical trend of research in the field.

\end{document}